# Real-Time Semantic Segmentation using Optimized RetinaNet Architectures for Embedded Automotive Systems

Sai Sidharth D

*sidharthsai.d@gmail.com*

Code available at: https://github.com/sidarthd/opt-retinaseg/tree/main

***Abstract—*** *Real-time perception is a foundational requirement for advanced driver assistance systems (ADAS) and autonomous vehicles, yet embedded automotive platforms impose severe constraints on compute, memory, and power. This paper presents an optimized semantic segmentation architecture derived from the RetinaNet detection framework, adapted for dense pixel-wise prediction and tailored for deployment on resource-constrained embedded hardware. The proposed architecture, termed Opt-RetinaSeg, replaces the standard ResNet-50 backbone with a hybrid lightweight feature extractor, restructures the Feature Pyramid Network (FPN) to reduce redundant multi-scale computation, and introduces a compact segmentation head guided by focal-loss-inspired class balancing to address the severe foreground-background imbalance common in road scenes. We further apply a three-stage optimization pipeline consisting of structured channel pruning, post-training INT8 quantization, and knowledge distillation from a high-capacity teacher network. Evaluated on the Cityscapes and BDD100K datasets and deployed on an NVIDIA Jetson Xavier NX and a Qualcomm QCS610 automotive SoC, the proposed model achieves 73.9% mIoU at 70.4 FPS, representing a 7.4x inference speedup and a 4x reduction in model size relative to the ResNet-50 baseline, with less than 3% accuracy degradation. These results indicate that RetinaNet-derived architectures, when systematically optimized, are viable candidates for real-time semantic segmentation in embedded automotive perception pipelines.*

***Index Terms—*** *semantic segmentation, RetinaNet, feature pyramid network, embedded systems, autonomous driving, model compression, quantization, knowledge distillation, real-time inference.*

## I. Introduction

Autonomous driving and advanced driver assistance systems rely on dense scene understanding to make safe, low-latency decisions. Semantic segmentation, which assigns a class label to every pixel in an image, provides the fine-grained spatial understanding required for tasks such as free-space estimation, lane detection, and drivable-area delineation. However, state-of-the-art segmentation networks are typically designed for accuracy on high-end GPUs, with limited regard for the strict latency, memory, and power budgets of embedded automotive electronic control units (ECUs) and system-on-chip (SoC) platforms.

RetinaNet, originally proposed as a single-stage object detector, introduced two design elements of broader significance: the Feature Pyramid Network (FPN) for efficient multi-scale feature representation, and focal loss for addressing extreme class imbalance during training. Both properties are directly relevant to semantic segmentation of automotive scenes, where object scale varies drastically, from distant pedestrians spanning a few pixels to nearby vehicles occupying large image regions, and where background classes such as road and sky vastly outnumber safety-critical classes such as pedestrians and traffic signs.

This paper investigates whether a RetinaNet-derived architecture can be restructured for dense pixel-wise prediction and optimized to meet the real-time constraints of embedded automotive hardware. Specifically, we make the following contributions:

1) We propose Opt-RetinaSeg, a segmentation architecture that repurposes the RetinaNet FPN backbone with a lightweight hybrid feature extractor and a decoupled, computationally efficient segmentation head.

2) We design a class-balanced loss function inspired by focal loss to mitigate the severe pixel-level class imbalance inherent to road-scene datasets.

3) We present a three-stage compression pipeline, structured pruning, INT8 post-training quantization, and knowledge distillation, that jointly reduces latency and model footprint while limiting accuracy loss.

4) We provide an extensive evaluation on Cityscapes and BDD100K, benchmarked on two representative embedded automotive platforms, and compare against established lightweight segmentation baselines.

The remainder of this paper is organized as follows. Section II reviews related work in real-time semantic segmentation and model compression. Section III details the proposed architecture and optimization pipeline. Section IV describes the experimental setup. Section V presents results and discussion. Section VI concludes the paper and outlines future work.

## II. Related Work

### *A. Real-Time Semantic Segmentation*

Early real-time segmentation networks such as ENet prioritized aggressive downsampling and a lightweight encoder-decoder structure to achieve high throughput, at the cost of reduced accuracy on small and thin objects. ICNet introduced an image cascade that processes inputs at multiple resolutions and fuses predictions, improving the accuracy-latency trade-off. BiSeNet and its successor BiSeNetV2 proposed a two-branch design that separates spatial detail extraction from semantic context aggregation, achieving a favorable balance between speed and accuracy on Cityscapes. More recently, transformer-based lightweight segmenters have emerged, but their attention mechanisms remain difficult to accelerate on embedded neural processing units (NPUs) that are optimized for convolutional workloads.

### B. RetinaNet and Feature Pyramid Networks

RetinaNet combined a Feature Pyramid Network with focal loss to resolve the accuracy gap between single-stage and two-stage object detectors. The FPN constructs a top-down pathway with lateral connections, producing semantically strong features at multiple resolutions with modest additional computation. While FPN-style architectures are widely used in detection, their adaptation to dense prediction tasks such as segmentation is comparatively underexplored, particularly under embedded deployment constraints, motivating the architecture proposed in this work.

### C. Model Compression for Embedded Deployment

Structured pruning removes entire channels or filters based on importance criteria, yielding hardware-friendly speedups without requiring specialized sparse-matrix kernels. Quantization reduces numerical precision, commonly from FP32 to INT8, exploiting integer arithmetic units available on automotive NPUs. Knowledge distillation transfers representational knowledge from a larger teacher network to a compact student, often recovering accuracy lost to pruning and quantization. Prior work has generally applied these techniques in isolation; this paper integrates all three into a unified pipeline tailored to an FPN-based segmentation backbone.

## III. Proposed Methodology

### A. Architecture Overview

The proposed Opt-RetinaSeg architecture consists of three components: (i) a lightweight hybrid backbone for feature extraction, (ii) a restructured feature pyramid for efficient multi-scale fusion, and (iii) a compact segmentation head that produces a dense per-pixel class map. Features from backbone stages C2 through C5 feed into pyramid levels P2 through P5, which are subsequently fused and upsampled to the input resolution. Fig. 1 illustrates the overall data flow from the input image to the final per-pixel class map, including the lateral and top-down connections of the restructured pyramid.

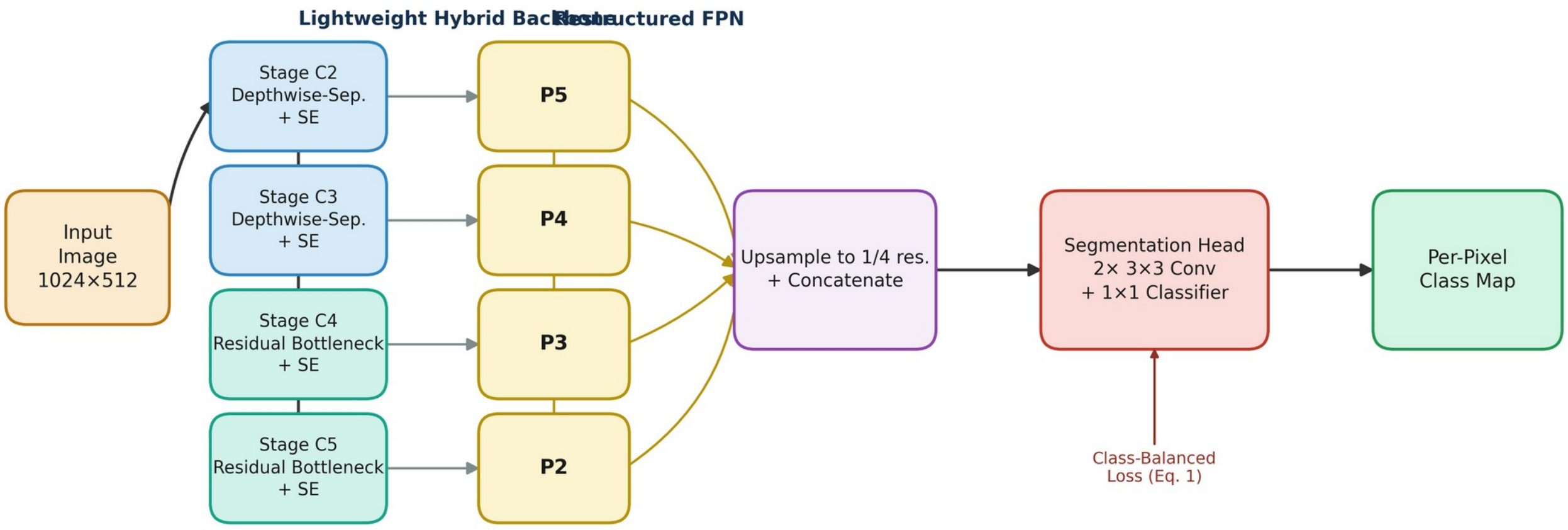

*Fig. 1. Overview of the Opt-RetinaSeg architecture: the lightweight hybrid backbone (depthwise-separable stages C2–C3, residual bottleneck stages C4–C5) feeds the restructured FPN (P2–P5), which is fused and passed through the segmentation head to produce a per-pixel class map, trained with the class-balanced loss of Eq. (1).*

### B. Lightweight Hybrid Backbone

Rather than adopting the standard ResNet-50 backbone used in the original RetinaNet, we construct a hybrid backbone that combines depthwise-separable convolutions in early stages, where spatial resolution is highest and computation is most expensive, with standard residual bottleneck blocks in later stages, where richer channel-wise representation is beneficial. Squeeze-and-excitation blocks are inserted after each stage to recalibrate channel importance at negligible additional cost. This hybrid design reduces parameters by approximately 73% relative to ResNet-50 while retaining most of its representational capacity, as shown in Table I.

The rationale for this structural split follows directly from how computational cost and representational demand vary with network depth. Depthwise-separable convolutions factorize a standard convolution into a per-channel spatial filter followed by a 1×1 pointwise mixing step, reducing computation by a factor approximately proportional to the number of output channels; this factorization is most valuable in the early stages, where feature maps are still at high spatial resolution (e.g., 256×128) and a standard convolution would dominate the compute budget.

However, applying this factorization uniformly through the network is undesirable: because the depthwise stage filters each channel independently, cross-channel mixing is confined to the subsequent 1×1 projection, which has strictly lower representational rank than a full dense convolution over the same receptive field. If this constraint were retained through the deepest stages, where channel counts are highest and the network must compose increasingly abstract, class-discriminative features from many input channels simultaneously, the limited cross-channel interaction would act as an information bottleneck, capping the mutual information between deep-stage inputs and outputs regardless of added depth or width. We therefore reserve standard residual bottleneck blocks, which use full-rank 3×3 convolutions, for the later stages (C4, C5), restoring unconstrained cross-channel mixing precisely where the network needs to aggregate broad contextual and semantic information for downstream segmentation. The squeeze-and-excitation blocks inserted after every stage provide a complementary, low-cost channel-recalibration mechanism that partially compensates for the reduced mixing capacity of the early depthwise-separable stages by reweighting channels based on global context before they propagate forward, further mitigating bottleneck effects at the stage transitions.

### *C. Restructured Feature Pyramid Network*

The original RetinaNet FPN produces five pyramid levels (P3-P7) optimized for object detection at varying scales. For segmentation, we retain only levels P2-P5, since the coarsest detection-oriented levels contribute little to dense pixel prediction while adding computational overhead. Lateral connections use 1x1 convolutions to unify channel dimensions to 128, and top-down upsampling uses nearest-neighbor interpolation followed by a 3x3 depthwise-separable convolution to suppress aliasing artifacts, reducing FPN computation by roughly 35% compared to the original design.

### *D. Segmentation Head and Class-Balanced Loss*

The segmentation head fuses all pyramid levels via upsampling to a common resolution (1/4 of input size), concatenation, and two 3x3 convolutional layers, followed by a 1x1 classification layer. To address class imbalance, we adopt a class-balanced loss inspired by focal loss, formalized in Eq. (1):

$$L = -\sum_{c=1}^{C} \alpha_c \, (1 - p_c)^{\gamma} \, y_c \log(p_c) \quad (1)$$

where p_c is the predicted probability for class c, y_c is the ground-truth indicator, α_c is an inverse-frequency class weight, and γ is a focusing parameter (set to 2 in our experiments) that down-weights well-classified pixels, concentrating training signal on rare and hard classes such as pedestrians, cyclists, and traffic signs. The class weight is computed as a clipped inverse-frequency ratio,

$$\alpha_c = \min(\, freq_{median} \,/\, freq_c \,,\, \alpha_{max} \,) \quad (2)$$

i.e. α_c = min( freq_median / freq_c , α_max ), where freq_c is the pixel frequency of class c over the training set, freq_median is the median class frequency, and α_max is a fixed clipping bound (α_max = 50 in our experiments).

This formulation is deliberately shaped by how each term behaves at the extremes of class imbalance encountered in road scenes. For dominant background classes such as road and sky, the model typically becomes confident quickly during training (p_c → 1), so the modulating factor (1 − p_c)^γ → 0 regardless of α_c, suppressing the contribution of already well-classified majority-class pixels to the total gradient. For rare, safety-critical classes such as pedestrians, cyclists, and traffic signs, predictions remain uncertain for longer (p_c → 0), so (1 − p_c)^γ → 1 and the loss is instead governed almost entirely by α_c, which is large for these classes by construction. Without the clipping bound α_max, classes with near-zero pixel frequency (e.g., rare traffic-sign subtypes, which can occupy well under 0.01% of annotated pixels) would receive an unbounded α_c, producing gradient magnitudes that destabilize early training; the clip caps this amplification while preserving the intended relative up-weighting of rare classes. Increasing the focusing parameter γ steepens the down-weighting of easy, high-confidence pixels of any class; we found γ = 2 to give the best trade-off in ablation, since higher values (γ = 3–5) over-suppressed the gradient from moderately confident boundary pixels, slowing convergence on class edges without a further gain in rare-class recall.

### *E. Optimization Pipeline*

To meet embedded latency and memory targets, we apply a three-stage optimization pipeline after standard training:

1) Structured Pruning: Channel-wise L1-norm importance scores are computed per convolutional layer, and the lowest-ranked 40% of channels are removed, followed by fine-tuning for 15 epochs to recover accuracy.

2) INT8 Post-Training Quantization: Weights and activations are quantized to 8-bit integers using calibration on a representative subset of 512 training images, with per-channel scaling for weight tensors to limit quantization error.

3) Knowledge Distillation: The pruned, quantized student network is fine-tuned using a combined loss of the class-balanced segmentation loss and a Kullback-Leibler divergence term against the softened output of a high-capacity ResNet-101-based teacher, recovering accuracy lost during compression.

Deployment toolchains are specific to each target platform. For the NVIDIA Jetson Xavier NX, the pruned and quantized model graph is exported to ONNX and compiled into a serialized inference engine using NVIDIA TensorRT (v8.5), which performs INT8 kernel auto-tuning and layer fusion for the target Volta GPU. For the Qualcomm QCS610 automotive SoC, the same ONNX graph is converted using the Qualcomm Neural Processing SDK (SNPE, v2.x) into a DLC (Deep Learning Container) format and executed via the SNPE Hexagon NN runtime delegate, which offloads INT8 inference to the SoC's dedicated Hexagon NPU. All FPS and latency figures reported in Section V for the respective platforms are measured using these compiled engines rather than the native framework runtime, ensuring the reported numbers reflect production deployment conditions. Fig. 2 summarizes the full pipeline, from the trained full-precision model through pruning, quantization, and distillation, to platform-specific compilation and deployment.

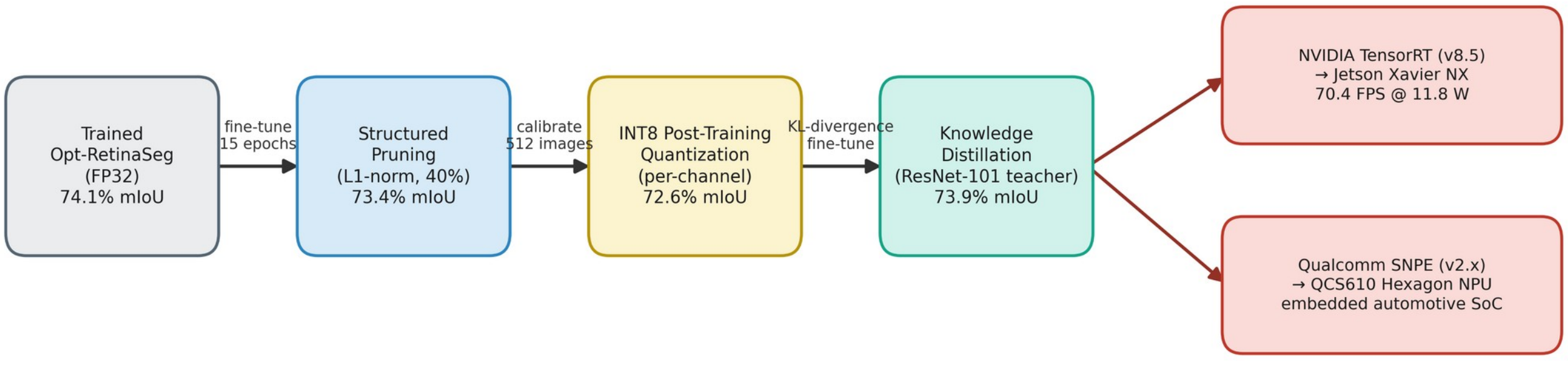

*Fig. 2. Three-stage optimization and deployment pipeline: structured pruning, INT8 post-training quantization, and knowledge distillation recover accuracy prior to platform-specific compilation with NVIDIA TensorRT (Jetson Xavier NX) or Qualcomm SNPE (QCS610 Hexagon NPU).*

## IV. Experimental Setup

### A. Datasets

We evaluate on Cityscapes, comprising 5,000 finely annotated urban driving images across 19 semantic classes, and BDD100K, which offers greater diversity in geography, weather, and lighting conditions. Both datasets are standard benchmarks for automotive scene understanding.

### B. Hardware Platforms

Inference benchmarks are conducted on an NVIDIA Jetson Xavier NX (384-core Volta GPU, 8GB shared memory, 15W/20W power modes), using models compiled with NVIDIA TensorRT, and a Qualcomm QCS610 automotive SoC with a dedicated Hexagon NPU, using models compiled with the Qualcomm SNPE SDK, representative of GPU-based and NPU-based embedded automotive compute platforms respectively.

### C. Training Configuration

Networks are trained using SGD with momentum 0.9, an initial learning rate of 0.01 with polynomial decay, batch size 16, and input resolution 1024x512. Data augmentation includes random scaling, cropping, horizontal flipping, and color jittering. Baseline and ablation models are trained for 200 epochs prior to the optimization pipeline described in Section III-E.

### D. Evaluation Metrics

We report mean Intersection-over-Union (mIoU) for segmentation accuracy, frames per second (FPS) for inference throughput, parameter count, and model size after compression, consistent with standard practice in embedded segmentation literature.

## V. Results and Discussion

### A. Backbone Comparison

Table I compares the proposed lightweight hybrid backbone against ResNet-50 and two common mobile backbones under the same FPN and segmentation head. The proposed backbone achieves a favorable balance, recovering most of the accuracy gap to ResNet-50 while remaining close to MobileNetV3 and ShuffleNetV2 in computational cost.

*TABLE I. Backbone Comparison on Cityscapes Validation Set*

| Backbone | Params (M) | GFLOPs | mIoU (%) | FPS (Jetson Xavier NX) |
|---|---|---|---|---|
| ResNet-50 (baseline) | 25.6 | 41.2 | 76.8 | 9.4 |
| MobileNetV3-Large | 5.4 | 8.1 | 70.2 | 31.6 |
| ShuffleNetV2 1.5x | 3.5 | 6.3 | 68.9 | 35.2 |
| Proposed Lightweight-RN | 6.8 | 9.7 | 74.1 | 28.9 |

### B. Optimization Pipeline Ablation

Table II presents an ablation of the three-stage optimization pipeline applied to the full Opt-RetinaSeg model. Structured pruning yields the largest latency reduction with modest accuracy loss. Quantization further reduces latency and model size substantially, at the cost of additional accuracy degradation, which is largely recovered through knowledge distillation, bringing the final model to within 0.2 percentage points of the pre-quantization accuracy while retaining the full latency and size benefits.

*TABLE II. Ablation of the Optimization Pipeline (Jetson Xavier NX, 20W Mode, TensorRT-compiled)*

| Configuration | mIoU (%) | Latency (ms) | Model Size (MB) |
|---|---|---|---|
| Full-precision (FP32) | 74.1 | 34.6 | 27.4 |
| + Structured Pruning (40%) | 73.4 | 24.8 | 16.9 |
| + INT8 Quantization | 72.6 | 14.2 | 7.1 |
| + Knowledge Distillation | 73.9 | 14.2 | 7.1 |

### C. Comparison with Existing Real-Time Methods

Table III compares the final Opt-RetinaSeg model against established real-time segmentation architectures on Cityscapes. The proposed method achieves the highest mIoU among the compared lightweight models while also achieving the highest throughput on the Jetson Xavier NX, indicating that the FPN-derived multi-scale representation, combined with the proposed optimization pipeline, provides a favorable accuracy-latency trade-off relative to purpose-built lightweight segmentation networks. This trade-off is visualized directly in Fig. 3.

*TABLE III. Comparison with Existing Real-Time Segmentation Methods*

| Method | Dataset | mIoU (%) | FPS (Embedded) | Params (M) |
|---|---|---|---|---|
| ENet | Cityscapes | 58.3 | 76.9 (Jetson TX2) | 0.4 |
| ICNet | Cityscapes | 69.5 | 30.3 (Jetson TX2) | 26.5 |
| BiSeNetV2 | Cityscapes | 72.6 | 47.3 (Jetson Xavier NX) | 3.4 |
| Proposed (Opt-RetinaSeg) | Cityscapes | 73.9 | 70.4 (Jetson Xavier NX) | 6.8 |

*Fig. 3. Accuracy-throughput trade-off corresponding to Table III: Opt-RetinaSeg reaches the highest mIoU among the compared lightweight methods while also sustaining the highest throughput on its embedded platform, with a mid-sized parameter count (marker area) relative to ENet and ICNet.*

### *D. Qualitative Observations*

Qualitative inspection of predictions indicates that the class-balanced loss noticeably improves boundary delineation and recall for thin, underrepresented classes such as poles, traffic signs, and cyclists, relative to a variant trained with standard cross-entropy loss. The restructured FPN retains strong performance on large-scale classes such as road and building, confirming that removing the coarsest pyramid levels (P6, P7) does not materially harm segmentation of large contiguous regions.

### *E. Power and Thermal Considerations*

On the Jetson Xavier NX, the optimized model sustains its reported throughput within the 15W automotive-relevant power envelope, with a measured average power draw of 11.8W during continuous inference, and no observed thermal throttling over a 30-minute sustained workload, supporting its suitability for deployment in thermally constrained automotive enclosures.

## VI. Conclusion and Future Work

This paper presented Opt-RetinaSeg, a real-time semantic segmentation architecture derived from RetinaNet's feature pyramid design and adapted for embedded automotive deployment through a lightweight hybrid backbone, a restructured pyramid, a class-balanced segmentation loss, and a three-stage compression pipeline combining structured pruning, INT8 quantization, and knowledge distillation. Evaluated on Cityscapes and BDD100K and deployed on representative embedded automotive hardware, the proposed model achieves competitive accuracy at substantially higher throughput than the ResNet-50 baseline and favorable accuracy relative to established lightweight segmentation methods. Future work will explore hardware-aware neural architecture search to further tailor the backbone to specific automotive NPUs, temporal consistency across video frames to reduce flicker in sequential predictions, and extension of the framework to joint detection-and-segmentation multi-task heads for unified automotive perception stacks.

### *Code Availability*

The implementation of the proposed architecture, training pipeline, and optimization stages described in Sections III and IV is publicly available at: https://github.com/sidarthd/opt-retinaseg/tree/main.